\documentclass[runningheads]{llncs}
\usepackage{graphicx}
\usepackage{amsmath,amssymb} %
\usepackage{color}
\usepackage[width=122mm,left=12mm,paperwidth=146mm,height=193mm,top=12mm,paperheight=217mm]{geometry}

\usepackage{caption}
\usepackage{booktabs}
\usepackage{tabularx}
\usepackage{subfig}
\usepackage{xspace}
\usepackage[normalem]{ulem}
\usepackage{url}
\usepackage{array}

\begin{document}
 \makeatletter
 \DeclareRobustCommand\onedot{\futurelet\@let@token\@onedot}
 \def\@onedot{\ifx\@let@token.\else.\null\fi\xspace}
 \def\eg{e.g\onedot} \def\Eg{E.g\onedot}
 \def\ie{i.e\onedot} \def\Ie{I.e\onedot}
 \def\cf{cf\onedot} \def\Cf{Cf\onedot}
 \def\etc{etc\onedot} \def\vs{vs\onedot}
 \def\wrt{w.r.t\onedot} \def\dof{d.o.f\onedot}
 \def\etal{\textit{et~al\onedot}} \def\iid{i.i.d\onedot}
 \def\Fig{Fig\onedot} \def\Eqn{Eqn\onedot} \def\Sec{Sec\onedot}
 \def\vs{vs\onedot}
 \makeatother

\DeclareRobustCommand{\figref}[1]{Fig.~\ref{#1}}
\DeclareRobustCommand{\figsref}[1]{Figures~\ref{#1}}

\DeclareRobustCommand{\Figref}[1]{Fig.~\ref{#1}}
\DeclareRobustCommand{\Figsref}[1]{Figures~\ref{#1}}

\DeclareRobustCommand{\Secref}[1]{Section~\ref{#1}}
\DeclareRobustCommand{\secref}[1]{Section~\ref{#1}}

\DeclareRobustCommand{\Secsref}[1]{Sections~\ref{#1}}
\DeclareRobustCommand{\secsref}[1]{Sections~\ref{#1}}

\DeclareRobustCommand{\Tableref}[1]{Table~\ref{#1}}
\DeclareRobustCommand{\tableref}[1]{Table~\ref{#1}}

\DeclareRobustCommand{\Tablesref}[1]{Tables~\ref{#1}}
\DeclareRobustCommand{\tablesref}[1]{Tables~\ref{#1}}

\DeclareRobustCommand{\eqnref}[1]{Equation~(\ref{#1})}
\DeclareRobustCommand{\Eqnref}[1]{Equation~(\ref{#1})}

\DeclareRobustCommand{\eqnsref}[1]{Equations~(\ref{#1})}
\DeclareRobustCommand{\Eqnsref}[1]{Equations~(\ref{#1})}

\DeclareRobustCommand{\chapref}[1]{Chapter~\ref{#1}}
\DeclareRobustCommand{\Chapref}[1]{Chapter~\ref{#1}}

\DeclareRobustCommand{\chapsref}[1]{Chapters~\ref{#1}}
\DeclareRobustCommand{\Chapsref}[1]{Chapters~\ref{#1}}

\hyphenation{po-si-tive}
\newcommand{\scream}[1]{\textbf{*** #1! ***}}
 \newcommand{\fixme}[1]{\textcolor{red}{\textbf{FiXme}#1}\xspace}
 \newcommand{\hobs}{\textrm{h}_\textrm{obs}}
 \newcommand{\cpad}[1]{@{\hspace{#1mm}}}
 \newcommand{\alg}[1]{\textsc{#1}}

 \newcommand{\fnrot}[2]{\scriptsize\rotatebox{90}{\begin{minipage}{#1}\flushleft #2\end{minipage}}}
 \newcommand{\chmrk}{{\centering\ding{51}}}
 \newcommand{\eqn}[1]{\begin{eqnarray}#1\end{eqnarray}}
 \newcommand{\eqns}[1]{\begin{eqnarray*}#1\end{eqnarray*}}

\newcommand{\todo}[1]{\textcolor{red}{ToDo: #1}}
\newcommand{\myparagraph}[1]{\noindent \textbf{#1}}

\newcommand{\marcus}[1]{\textcolor{green}{Marcus: #1}}
\newcommand{\anja}[1]{\textcolor{red}{Anja: #1}}
\newcommand{\trevor}[1]{\textcolor{blue}{Trevor: #1}}

\newcommand{\invisible}[1]{}%

\newcommand{\figvspace}{}
\newcommand{\secvspace}{}
\newcommand{\subsecvspace}{}

\newcommand{\approach}{GroundeR\xspace}

\graphicspath{{./fig/}{./fig/plots/}}

\pagestyle{headings}
\mainmatter

\title{Grounding of Textual Phrases in Images by Reconstruction} %

\titlerunning{Grounding of Textual Phrases in Images by Reconstruction}

\authorrunning{A. Rohrbach, M. Rohrbach, R.  Hu, T. Darrell, and B. Schiele}

 \newcommand{\authSpace}{\ \ \ \ }
 \author{
 Anna Rohrbach$^{1}$ \authSpace Marcus Rohrbach$^{2,3}$ \authSpace Ronghang  Hu$^{2}$ \\ Trevor Darrell$^{2}$ \authSpace Bernt Schiele$^{1}$
 }

\institute{
$^{1}$Max Planck Institute for Informatics, Saarbr{\"u}cken, Germany\\
$^{2}$UC Berkeley EECS,  CA, United States\\
$^{3}$ICSI, Berkeley, CA, United States\\
\email{ \{arohrbach,schiele\}@mpi-inf.mpg.de, \{rohrbach,ronghang,trevor\}@eecs.berkeley.edu}
}

\maketitle 
\begin{abstract}
Grounding (\ie localizing) arbitrary, free-form textual phrases in visual content is a challenging problem with many applications for human-computer interaction and image-text reference resolution. Few datasets provide the ground truth spatial localization of phrases, thus it is desirable to learn from data with no or little grounding supervision. We propose a novel approach which learns grounding by reconstructing a given phrase using an attention mechanism, which can be either latent or optimized directly. During training our approach encodes the phrase using a recurrent network language model and then learns to attend to the relevant image region in order to reconstruct the input phrase. At test time, the correct attention, \ie, the grounding, is evaluated. If grounding supervision is available it can be directly applied via a loss over the attention mechanism. We demonstrate the effectiveness of our approach on the Flickr 30k Entities~\cite{plummer15iccv} and ReferItGame~\cite{kazemzadeh14emnlp} datasets with different levels of supervision, ranging from no supervision over partial supervision to full supervision. Our supervised variant improves by a large margin over the state-of-the-art on both datasets.  
\end{abstract}

\section{Introduction}

\begin{figure}[t]
\footnotesize
\begin{tabular}{p{0.33\textwidth}@{\ \ }p{0.30\textwidth}@{\ \ }p{0.25\textwidth}}
{\includegraphics[width=\linewidth]{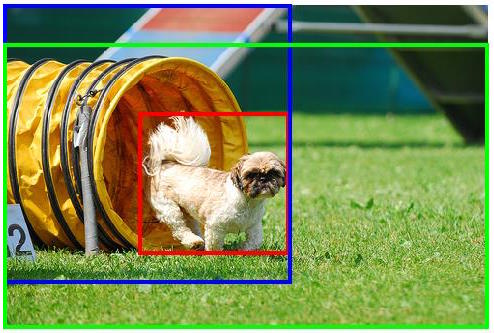}}
{\textcolor{red}{A little brown and white dog} emerges from \textcolor{blue}{a yellow collapsable toy tunnel} onto \textcolor{green}{the lawn}.} &
\raisebox{-75pt}
{\includegraphics[width=\linewidth]{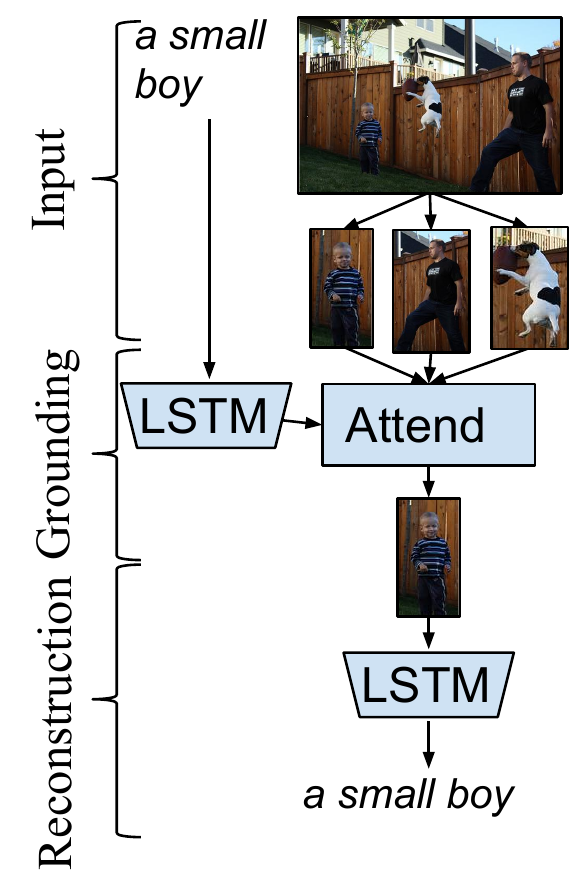}} & 
\raisebox{-35pt}
{\includegraphics[width=\linewidth]{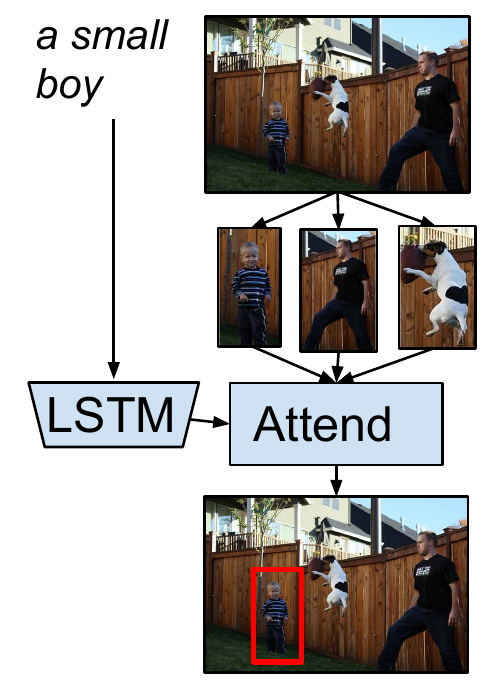}} \\
(a) Predicted grounding. &
(b) Training time. 
  &
(c) Test time. \\
\end{tabular}
\caption{(a) Without bounding box annotations at training time our approach \approach can ground free-form natural language phrases in images. %
(b) During training our latent attention approach reconstructs phrases by learning to attend to the correct box. (c) At test time, the attention model infers the grounding for each phrase. For semi-supervised and fully supervised variants see \Figref{fig:net}.}
\label{fig:concept}
\label{fig:teaser}
\end{figure}

Language grounding in visual data is an interesting problem studied both in computer vision \cite{karpathy15cvpr,karpathy14nips,kong14cvpr,plummer15iccv,hu16cvpr} and natural language processing \cite{krishnamurthy13tacl,matuszek12icml} communities. Such grounding can be done on different levels of granularity: from coarse, \eg associating a paragraph of text to a scene in a movie \cite{tapaswi2015cvpr,zhu2015aligning}, to fine, \eg localizing a word or phrase in a given image \cite{plummer15iccv,hu16cvpr}. In this work we focus on the latter scenario. Many prior efforts in this area have focused on rather constrained settings with a small number of nouns to ground \cite{lin14cvpr,kong14cvpr}. On the contrary, we want to tackle the problem of grounding  arbitrary natural language phrases in images. Most parallel corpora of sentence/visual data do not provide localization annotations (\eg bounding boxes) and the annotation process is costly. 
We propose an approach which can learn to localize phrases relying only on phrases associated with images without bounding box annotations but which is also able to incorporate phrases with bounding box supervision when available (see 
\Figref{fig:teaser}).

The main idea of our approach is shown in \Figref{fig:concept}(b,c). Let us first consider the scenario where no localization supervision is available. Given images paired with natural language phrases %
we want to localize these phrases with a bounding box in the image (\Figref{fig:concept}c). To do this we propose a model (\Figref{fig:concept}b) which learns to attend to a bounding box proposal and, based on the selected bounding box, reconstructs the phrase.  As the second part of the model (\Figref{fig:concept}b, bottom) is able to predict the correct phrase only if the first part of the model attended correctly (\Figref{fig:concept}b, top), this can be learned without additional bounding box supervision. %
Our method is  based on \emph{Ground}ing with a \emph{R}econstruction loss and hence named \emph{GroundeR}.
Additional supervision is integrated in our model by adding a loss function which directly penalizes incorrect attention before the reconstruction step. %
At test time we evaluate whether the model attends to the correct bounding box.

We propose a novel approach to grounding of textual phrases in images %
which can operate in all supervision modes: with no, a few, or all grounding annotations available. We evaluate our GroundeR approach on the Flickr 30k Entities \cite{plummer15iccv} and ReferItGame \cite{kazemzadeh14emnlp} datasets and show that our unsupervised variant is  better than prior work and our supervised approach significantly outperforms state-of-the-art on both datasets. Interestingly, our semi-supervised approach can effectively exploit small amounts of labeled data and surpasses the supervised variant by exploiting multiple losses.

\section{Related work}
\label{sec:related}
\myparagraph{Grounding natural language in images and video.}
For grounding language in images, the approach of \cite{kong14cvpr} is based on a Markov Random Field which aligns 3D cuboids to words. However it is limited to nouns of 21 object classes relevant to indoor scenes. \cite{johnson2015cvpr} uses a Conditional Random Field to ground the specifically designed scene graph query in the image. 
\cite{karpathy14nips} grounds dependency-tree relations to image regions using Multiple Instance Learning and a ranking objective. \cite{karpathy15cvpr} simplifies this objective to just the maximum score and replaces the dependency tree with a learned recurrent network. Both works have not been evaluated for grounding, but we discuss a quantitative comparison in \secref{sec:results}.
Recently, \cite{plummer15iccv} presented a new dataset, Flickr 30k Entities, which augments the Flickr30k dataset \cite{young2014image} with bounding boxes for all noun phrases present in textual descriptions. \cite{plummer15iccv} report the localization performance of their proposed CCA embedding \cite{gong2014eccv} approach. \cite{wang2016cvpr} proposes Deep Structure-Preserving Embedding for image-sentence retrieval and also applies it to phrase localization, formulated as ranking problem. The Spatial Context Recurrent ConvNet (SCRC) \cite{hu16cvpr} and the approach of \cite{mao16cvpr} use a caption generation framework to score the phrase on the set of proposal boxes, to select the box with highest probability. %
One advantage of our approach over \cite{hu16cvpr,mao16cvpr} is its applicability to un- and semi-supervised training regimes. We believe that our approach of encoding the phrase optimizes the better objective for grounding than scoring the phrase with a text generation pipeline as in \cite{hu16cvpr,mao16cvpr}. As for the fully-supervised regime we empirically show our advantage over \cite{hu16cvpr}. \cite{sadeghi2015viske} attempts to localize relation phases of type Subject-Verb-Object at a large scale in order to verify their correctness, while relying on detectors from \cite{divvala2014learning}.

In the video domain some of the representative works on spatial-temporal language grounding are \cite{lin14cvpr} and \cite{yu2013acl}. These are limited to small set of nouns.

\myparagraph{Object co-localization}
focuses on discovering and detecting an object in images or videos without any bounding box annotation, but only from image/video level labels \cite{blaschko2010simultaneous,cinbis2014multi,joulin2014eccv,kwak15arxiv,song2014learning,tang2014cvpr,yu15arxiv}. %
These works are similar to ours with respect to the amount of supervision, but they focus on a few discrete classes, while our approach can handle arbitrary phrases and allows for localization of novel phrases. There are also works that propose to train detectors for a wide range of concepts using image-level annotated data from web image search, \eg \cite{divvala2014learning} and \cite{chen2015webly}. These approaches are complementary to ours in the sense of obtaining large scale concept detectors with little supervision, however they do not tackle complex phrases \eg ``a blond boy on the left'' which is the focus of our work.

\myparagraph{Attention in vision tasks.}
Recently, different attention mechanisms have been applied to a range of computer vision tasks. The general idea is that given a visual input, \eg set of features, at any given moment we might want to focus only on part of it, \eg attend to a specific subset of features \cite{bahdanau2014neural}. \cite{xu2015arxiv} integrates spatial attention into their image captioning pipeline. They consider two variants: ``soft'' and ``hard'' attention, meaning that in the latter case the model is only allowed to pick a single location, while in the first one the attention ``weights'' can be distributed over multiple locations.  \cite{jin2015aligning} adapts the soft-attention mechanism and attends to bounding box proposals, one word at a time, while generating an image captioning. \cite{yao2015iccv} relies on a similar mechanism to perform temporal attention for selecting frames in video description task. \cite{yeung2015every} uses attention mechanism to densely label actions in a video sequence. Our approach relies on soft-attention mechanism, similar to the one of \cite{xu2015arxiv}. We apply it to the language grounding task where attention helps us to select a bounding box proposal for a given phrase.

\myparagraph{Bi-directional mapping.}
In our model, a phrase is first mapped to a image region through attention, and then the image region is mapped back to phrase during reconstruction. There is conceptual similarity between previous work and ours on the idea of bi-directional mapping from one domain to another. In autoencoders \cite{vincent2008extracting}, input data is first mapped to a compressed vector during encoding, and then reconstructed during decoding. \cite{chen15cvpr} uses a bi-directional mapping from visual features to words and from words to visual features in a recurrent neural network model. The idea is to generate descriptions from visual features and then to reconstruct visual features given a description. Similar to \cite{chen15cvpr}, our model can also learn to associate input text with visual features, but through attending to an image region rather than reconstructing directly from words. 
In the linguistic community, \cite{ammar2014conditional} proposed a CRF Autoencoder, which generates latent structures for the given language input and then reconstructs the input from these latent structures, with the application to \eg part-of-speech tagging.

\section{\approach: \emph{Ground}ing by \emph{R}econstruction}

The goal of our approach is to ground natural language phrases in images. More specifically, to ground a phrase $p$ in an image $I$ means to find a region $r_j$ in the image which corresponds to this phrase. $r_j$ can be any subset of $I$, \eg a segment or a bounding box. The core insight of our method is that there is a bi-directional correspondence between an image region and the phrase describing it. As a correct grounding of a textual phrase should result in an image region which a human would describe using this phrase, \ie it is possible to reconstruct the phrase based on the grounded image region. Thus, the key idea of our approach is to learn to ground a phrase by reconstructing this phrase from an automatically localized region. \Figref{fig:concept} gives an overview of our approach.

\begin{figure}[t]
\footnotesize
\begin{tabular}{p{0.33\textwidth}@{\ \ }p{0.33\textwidth}@{\ \ }p{0.33\textwidth}}
\includegraphics[width=\linewidth]{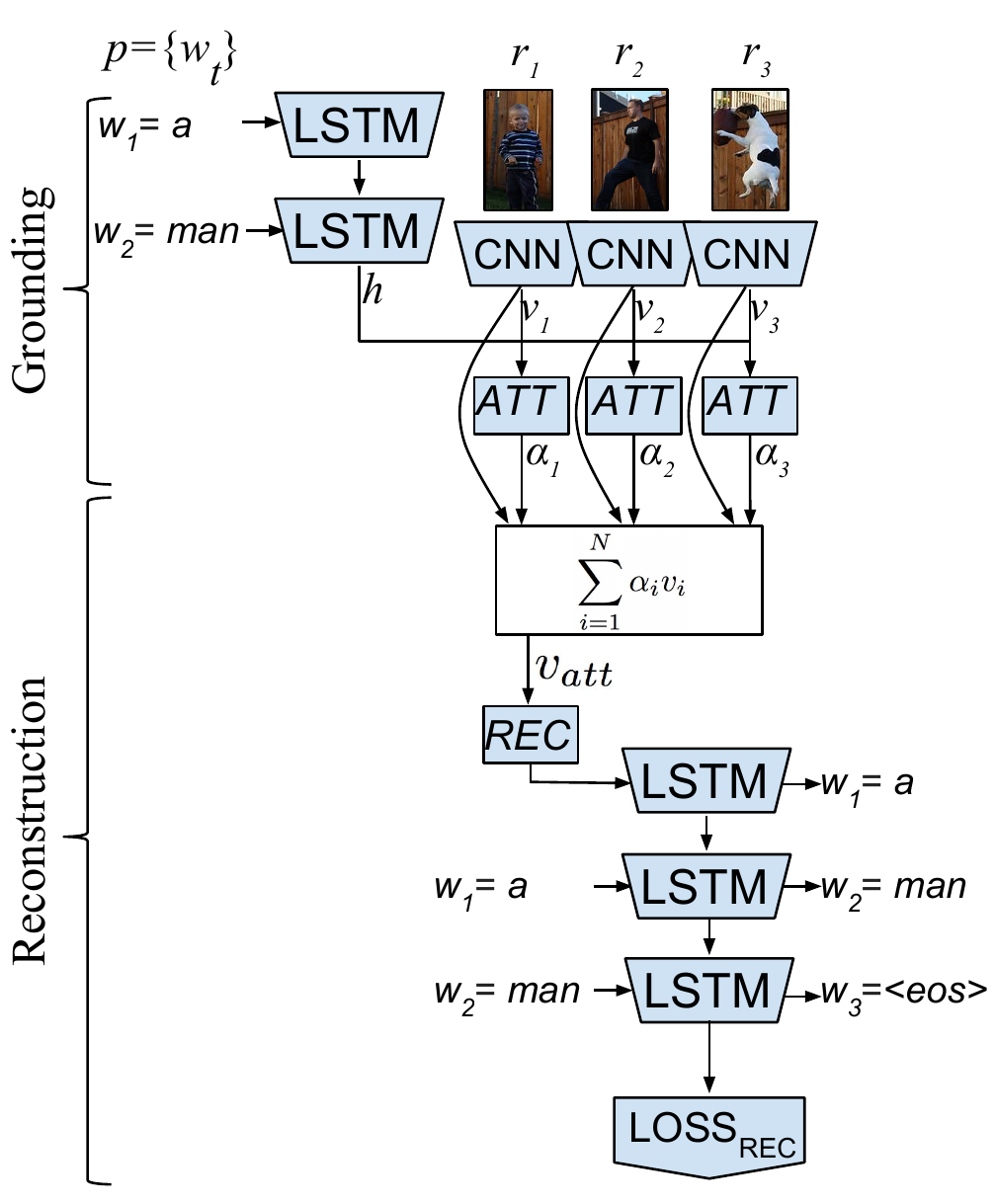}&
\includegraphics[width=\linewidth]{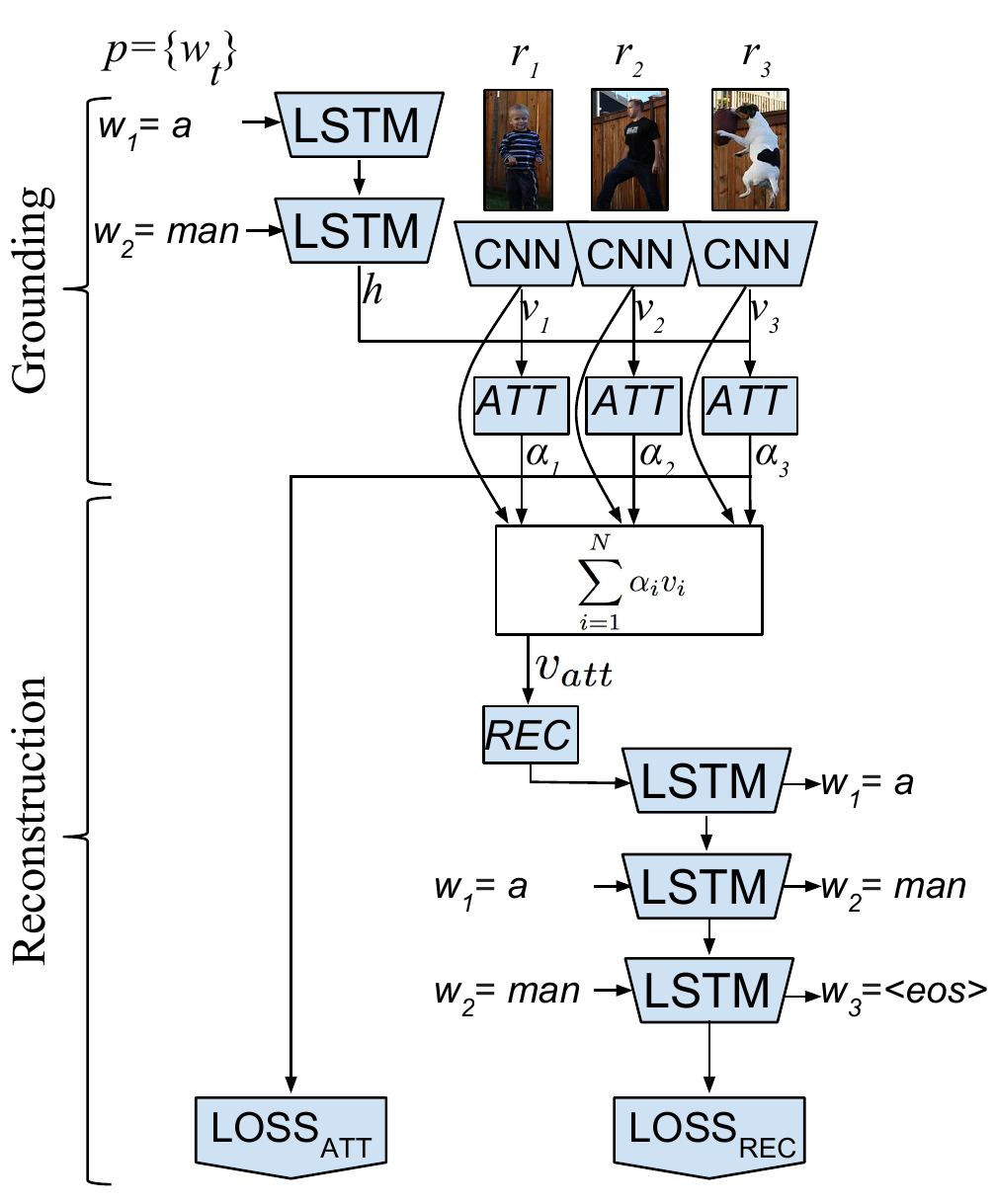} & 
\includegraphics[width=\linewidth]{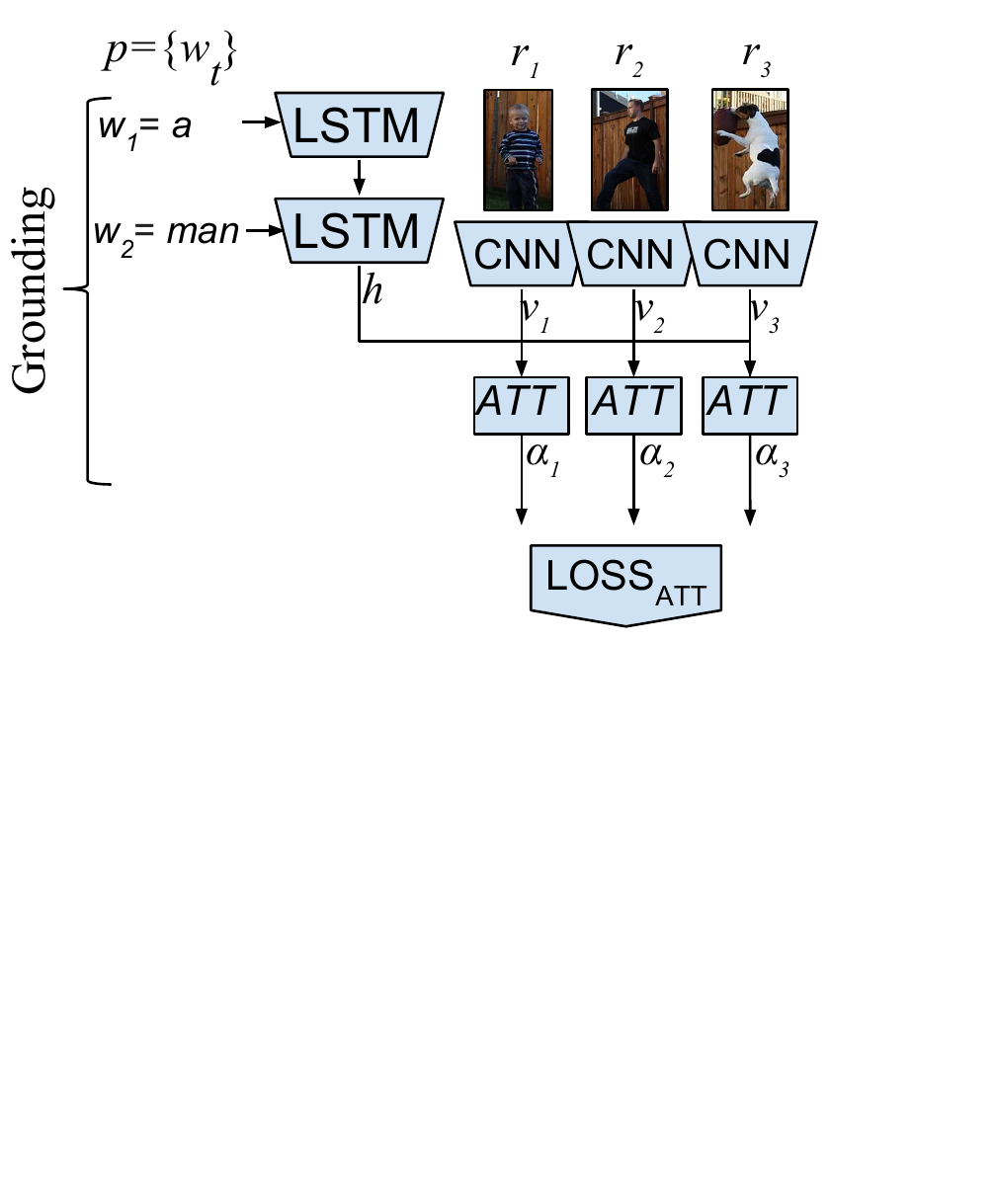}\\
(a) Unsupervised & 
(b) Semi-supervised &
(c) Fully supervised \\
\end{tabular}
\caption{Our model learns grounding of textual phrases in images with (a) no, (b) little  (c)  or full supervision of localization, through a grounding part and a reconstruction part. During training, the model distributes its attention to a single or several boxes, and learns to reconstruct the input phrase based on the boxes it attends to. At test time, only the grounding part is used. %
}
\label{fig:net}
\end{figure}

In this work, we utilize a set of automatically generated bounding box proposals $\{r_i\}_{i\in N}$ for the image $I$. Given a phrase $p$, during training our model works in two parts: the first part aims to attend to the most relevant region $r_j$ (or potentially also multiple regions) based on the phrase $p$, and then the second part tries to reconstruct the same phrase $p$ from region(s) $r_j$ it attended to in the first phase. 
Therefore, by training to reconstruct the text phrase, the model learns to first ground the phrase in the image, and then generate the phrase from that region. \Figref{fig:net}a visualizes the network structure. At test time, we remove the phrase reconstruction part, and use the first part for phrase grounding. 
The described pipeline can be extended to accommodate partial supervision, \ie ground-truth phrase localization. For that we integrate an additional loss into the model, which directly optimizes for correct attention prediction, see \Figref{fig:net}b. Finally, we can adapt our model to the fully supervised scenario by removing the reconstruction phase, see \Figref{fig:net}c.

In the following we present the details of the two parts in our approach: learning to attend to the correct region for a given phrase and learning to reconstruct the phrase from the attended region. For simplicity, but without loss of generality, we will refer to $r_j$ as a single bounding box.

\subsection{Learning to ground}

We frame the problem of grounding a phrase $p$ in image $I$ as selecting a bounding box $r_j$ from a set of image region proposals $\{r_i\}_{i=1,\cdots, N}$. To select the correct bounding box, we define an attention function $f_{ATT}$ and select the box $j$ which receives the maximum attention:
\begin{equation}
j = \arg\max_{i}f_{ATT}(p,r_i)
\end{equation} 

In the following we describe the details of how we model the attention in $f_{ATT}$.
The attention mechanism used in our model is inspired by and similar to the soft attention formulations of \cite{jin2015aligning,xu2015arxiv}. However, our inputs to the attention predictor are not single words but rather multi-word phrases, and consequently we also do not have a ``doubly stochastic attention'' which is used in \cite{xu2015arxiv} to normalize the attention across words.

The phrases that we are dealing with might be very complex thus we require a good language model to represent them. We choose a Long Short-Term Memory network (LSTM) \cite{hochreiter1997long} as our phrase encoder, as it has been shown effective in various language modeling tasks, \eg translation \cite{sutskever14nips}. We encode our query phrase word by word with an LSTM and obtain a representation of the phrase using the hidden state $h$ at the final time step as:
\begin{equation}
h=f_{LSTM}(p)
\end{equation}
Each word $w_t$ in the phrase $p$ is first encoded with a one-hot-vector.
Then it is embedded in the lower dimensional space and given to LSTM. 

Next, each bounding box $r_i$ is encoded using a convolutional neural network (CNN) to compute the visual feature vector $v_i$:
\begin{equation}
v_i=f_{CNN}(r_i)
\end{equation}

Based on the encoded phrase and feature representation of each proposal, we use a two layer perceptron to compute the attention on the proposal $r_i$: 
\begin{equation}
\bar{\alpha}_i = f_{ATT}(p, r_i)=W_2\phi(W_h h + W_v v_i + b_1)+b_2
\end{equation}
where $\phi$ is the rectified linear unit (ReLU): $\phi(x)=max(0,x)$. We found that this architecture performs better than \eg a single layer perceptron with a hyperbolic tangent nonlinearity used in \cite{bahdanau2014neural}.

We get normalized attention weights $\alpha_i$ by using softmax, which can be interpreted as probability of region $r_i$ being the correct region $r_{\hat{j}}$:
\begin{equation}
\label{eq:attention}
\alpha_i = P(i = \hat{j}| \bar{\alpha}) = \frac{\exp(\bar{\alpha}_i)}{\sum_{k=1}^N{\exp(\bar{\alpha}_k)}}
\end{equation}

 If at training time we have ground truth information, \ie that $r_{\hat{j}}$ is the correct proposal box, then we can compute the loss $L_{att}$ based on our prediction as: 
\begin{equation}
\label{eq:attloss}
L_{att} = -  \frac{1}{B} \sum_{b=1}^{B} \log(P({ \hat{j}}  | \bar{\alpha})),
\end{equation}
where $B$ is the number of phrases per batch.
This loss activates only if the training sample has the ground-truth attention value, otherwise, it is zero.
If we do not have ground truth annotations then we have to define a loss function to learn the parameters of $f_{ATT}$ in a weakly supervised manner. In the next section we describe how we define this loss by aiming to reconstruct the phrase based on the boxes that are attended to. At test time, we calculate the IOU (intersection over union) value between the selected box $r_j$ and the ground truth box $r_{\hat{j}}$.
\subsection{Learning to reconstruct}
\label{sec:reconstruct}
The key idea of our phrase reconstruction model is to learn to reconstruct the phrase only from the attended boxes. %
Given an attention distribution over the boxes, we compute a weighted sum over the visual features and the attention weights $\alpha_i$:
\begin{equation}
v_{att} = \sum_{i=1}^N{\alpha_i{v_i}},
\end{equation}
which aggregates the visual features from the attended boxes. Then, the visual features $v_{att}$ are further encoded into $v'_{att}$ using a non-linear encoding layer:
\begin{equation}
v'_{att}=f_{REC}(v_{att})=\phi(W_{a} v_{att} + b_a)
\end{equation}

We reconstruct the input phrase based on this encoded visual feature $v'_{att}$ over attended regions. During reconstruction, we use an image description LSTM that takes $v'_{att}$ as input to generate a distribution over phrases $p$:
\begin{equation}
P(p | v'_{att}) = f_{LSTM}(v'_{att})
\end{equation}
where $P(p | v'_{att})$ is a distribution over the phrases conditioned on the input visual feature.
Our approach for phrase generation is inspired by \cite{donahue15cvpr,vinyals14arxiv} who have effectively used LSTM for generating image descriptions based on visual features. Given a visual feature, it learns to predict a word sequence $\{w_t\}$. At each time step $t$, the model predicts a distribution over %
 the next word $w_{t+1}$ conditioned on the input visual feature $v'_{att}$ and all the previous words. %
 We use a single LSTM layer and we feed the visual input only at the first time step. We use LSTM as our phrase encoder as well as decoder. Although one could potentially use other approaches to map phrases into a lower dimensional semantic space, it is not clear how one would do the reconstruction without the  recurrent network, given that we have to train encoding and decoding end-to-end.

Importantly, the entire grounding+reconstruction model is trained as a single deep network through back-propagation by maximizing the likelihood of the ground truth phrase $\hat{p}$ generated during reconstruction, where we define the training loss for batch size $B$:
\begin{equation}
L_{rec} = - \frac{1}{B} \sum_{b=1}^{B} \log(P(\hat{p} | v'_{att}))
\end{equation}

Finally, in the semi-supervised model we have both losses $L_{att}$ and $L_{rec}$, which are combined as follows:
\begin{equation}
L = \lambda L_{att} + L_{rec}
\end{equation}
where parameter $\lambda$ regulates the importance of the attention loss.

\section{Experiments}
\label{sec:results}

We first discuss the experimental setup and design choices of our implementation and then present  quantitative results on the test sets of Flickr 30k Entities (\tablesref{tbl:testset_flickr},\ref{tbl:testset}) and ReferItGame (\tableref{tbl:testset_referit}) datasets. We find our best results to outperform state-of-the-art on both datasets by a significant margin. \Figsref{fig:qualitative_flickr} and \ref{fig:qualitative_referit} show qualitatively how well we can ground phrases in images.

\subsection{Experimental Setup}
\label{sec:expsetup}

We evaluate GroundeR on the datasets Flickr 30k Entities \cite{plummer15iccv} and ReferItGame \cite{kazemzadeh14emnlp}. Flickr 30k Entities \cite{plummer15iccv} contains over 275K bounding boxes from 31K images associated with natural language phrases. Some phrases in the dataset correspond to multiple boxes, \eg ``two men''. For consistency with \cite{plummer15iccv}, in such cases we consider the union of the boxes as ground truth. We use 1,000 images for validation, 1,000 for testing and 29,783 for training. 
The ReferItGame \cite{kazemzadeh14emnlp} dataset contains over 99K regions from 20K images. Regions are associated with natural language expressions, constructed to disambiguate the described objects. We use the bounding boxes provided by \cite{hu16cvpr} and the same test split, namely 10K images for testing; the rest we split in 9K training and 1K validation images.

We obtain 100 bounding box proposals for each image using Selective Search \cite{uijlings2013selective} for Flickr 30k Entities and Edge Boxes \cite{zitnick2014eccv} for ReferItGame dataset. For our semi-supervised and fully supervised models we obtain the ground-truth attention by selecting the proposal box which overlaps most with the ground-truth box, while the overlap IOU (intersection over union) is above 0.5. Thus, our fully supervised model is not trained with all available training phrase-box pairs, but only with those where such proposal boxes exist.

On the Flickr 30k Entities for the visual representation we rely on the VGG16 network \cite{simonyan2014very} trained on ImageNet \cite{deng09cvpr}. For each box we extract a 4,096 dimensional feature from the fully connected fc7 layer. We also consider a VGG16 network fine-tuned for object detection on PASCAL \cite{everingham2010pascal}, trained using Fast R-CNN \cite{girshick2015fast}. In the following we refer to both features as VGG-CLS and VGG-DET, respectively. We do not fine-tune the VGG representation for our task to reduce computational and memory load, however, our model trivially allows back-propagation into the image representation which likely would lead to further improvements. For the ReferItGame dataset we use the VGG-CLS features and additional spatial features provided by \cite{hu16cvpr}. We concatenate both and refer to the obtained feature as VGG+SPAT. For the language encoding and decoding we rely on the LSTM variant implemented in Caffe \cite{jia2014caffe} which we initialize randomly and jointly train with the grounding task.

At test time we compute the accuracy as the ratio of phrases for which the attended box overlaps with the ground-truth box by more than 0.5 IOU.

\subsection{Design choices and findings}

In all experiments we use the Adam solver \cite{kingma2014adam}, which adaptively changes the learning rate during training. We train our models for about 20/50 epochs for the Flickr 30k Entities/ReferItGame dataset, respectively, and pick the best iteration on the validation set. 

Next, we report our results for optimizing hyperparmeters on the validation set of Flickr 30k Entities while using the VGG-CLS features.

\myparagraph{Regularization.}
Applying L2 regularization to parameters (weight decay) is important for the best performance of our unsupervised model. By introducing the weight decay of $0.0005$ we improve the accuracy from $20.33\%$ to $22.96\%$. In contrast, when supervision is available, we introduce batch normalization \cite{ioffe2015batch} for the phrase encoding LSTM and visual feature, which leads to a performance improvement, in particular from 37.42\% to 40.93\% in the supervised scenario.

\myparagraph{Layer initialization.}
We experiment with different ways to initialize the layer parameters. The configuration which works best for us is using uniform initialization for LSTM, MSRA \cite{he2015delving} for convolutional layers, and Xavier \cite{glorot2010understanding} for all other layers. Switching from Xavier to MSRA initialization for the convolutional layers improves the accuracy of the unsupervised model from $21.04\%$ to $22.96\%$.%

\subsection{Experiments on Flickr 30k Entities dataset}
We report the performance of our approach with multiple levels of supervision in Table \ref{tbl:testset_flickr}. 
In the last line of the table we report the proposal upper-bound accuracy, namely the presence of the correct box among the proposals (which overlaps with the ground-truth box with $IOU > 0.5$).

\newcommand{\midruleValShort}{\cmidrule(rr){1-1} \cmidrule(rr){2-4}}
\begin{table*}[t]
\scriptsize
\center
\begin{tabular}{lccc}
\toprule
Approach & \multicolumn{3}{c}{Accuracy} \\
         & Other & VGG-CLS & VGG-DET \\
\midruleValShort
\multicolumn{4}{l}{\textbf{Unsupervised training}} \\
Deep Fragments [6] & 21.78 & - & - \\
GroundeR & - & 24.66 & 28.94 \\
\midruleValShort
\multicolumn{4}{l}{\textbf{Supervised training}} \\
CCA \cite{plummer15iccv} & - & 27.42 & - \\
SCRC \cite{hu16cvpr} & - & 27.80 & - \\
DSPE \cite{wang2016cvpr} & - & - & 43.89 \\
GroundeR & - & 41.56 & 47.81 \\
\midruleValShort
\multicolumn{4}{l}{\textbf{Semi-supervised training}} \\
GroundeR 3.12\% annot. & - & 33.02 & 42.32 \\
GroundeR 6.25\% annot. & - & 37.10 & 44.02 \\
GroundeR 12.5\% annot. & - & 38.67 & 44.96 \\
GroundeR 25.0\% annot. & - & 39.31 & 45.32 \\
GroundeR 50.0\% annot. & - & 40.72 & 46.65 \\
GroundeR 100.0\% annot. & - & 42.43 & 48.38 \\
\midruleValShort
Proposal upperbound & 77.90 & 77.90 & 77.90 \\
\bottomrule\\
\end{tabular}
\caption{Phrase localization performance on Flickr 30k Entities with different levels of bounding box supervision, accuracy in \%.}
\label{tbl:testset_flickr}
\end{table*}

\myparagraph{Unsupervised training.}
We start with the unsupervised scenario, \ie no phrase localization ground-truth is used at training time.
Our approach, which relies on VGG-CLS features, is able to achieve 24.66\% accuracy. Note that the VGG network trained on ImageNet has not seen any bounding box annotations at training time. VGG-DET, which was fine-tuned for detection, performs better and achieves 28.94\% accuracy. We can further improve this by taking a sentence constraint  into account. Namely, it is unlikely that two different phrases from one sentence are grounded to the same box. Thus we post-process the attended boxes: we jointly process the phrases from one sentence and greedily select the highest scoring box for each phrase, while the same box cannot be selected twice. This allows us to reach the accuracy of 25.01\% for VGG-CLS and 29.02\% for VGG-DET. While we currently only use a sentence constraint as a simple post processing step at test time, it would be interesting to include a sentence level constraint during training as part of future work. 
We compare to the unsupervised Deep Fragments approach of \cite{karpathy14nips}. 
Note, that \cite{karpathy14nips} does
not report the grounding performance and does not allow for direct comparison with our work. With our best case evaluation\footnote{%
We train the Deep Fragments model \cite{karpathy14nips} on the the Flickr 30k dataset and evaluate with the Flickr 30k Entities ground truth phrases and boxes. %
Our trained Deep Fragments model achieves 11.2\%/16.5\% recall@1 for image annotation/search compared to 10.3\%/16.4\% reported in \cite{karpathy14nips}. %
As there is a large number of dependency tree fragments per sentence (on average 9.5) which are matched to proposal boxes, rather than on average 3.0 noun phrases per sentence in Flickr 30k Entities, we make a best case study in favor of \cite{karpathy14nips}. For each ground-truth phrase we take the maximum overlapping dependency tree fragments (w.r.t. word overlap), compute the IOU between their matched boxes and the ground truth, and take the highest IOU.}
 of Deep Fragments \cite{karpathy14nips}, which also relies on detection boxes and features, we achieve an accuracy of 21.78\%. 
Overall, the ranking objective in \cite{karpathy14nips} can be seen complimentary to our reconstruction objective. It might be possible, as part of future work, to combine both objectives to learn even better models without grounding supervision.

\myparagraph{Supervised training.}
Next we look at the fully supervised scenario. The accuracy achieved by \cite{plummer15iccv} is 27.42\%\footnote{The number was provided by the authors of \cite{plummer15iccv}, while in \cite{plummer15iccv} they report 25.30\% for phrases automatically extracted with a parser.} and by  
SCRC \cite{hu16cvpr} is 27.80\%. Recent approach of \cite{wang2016cvpr} achieves 43.89\% with VGG-DET features. Our approach, when using VGG-CLS features achieves an accuracy of 41.56\%, significantly improving over prior works that use VGG-CLS. We further improve our result to impressive 47.81\% when using VGG-DET features.

\myparagraph{Semi-supervised training.}
Finally, we move to the semi-supervised scenario. The notation ``$x$\% annot.'' means that $x$\% of the annotated data (where ground-truth attention is available) %
is used. As described in Section \ref{sec:reconstruct} we have a parameter $\lambda$ which controls the weight of the attention loss $L_{att}$ vs. the reconstruction loss $L_{rec}$. We estimate the value of $\lambda$ on validation set and fix it for all iterations. We found that we need higher weight on $L_{att}$ when little supervision is available. E.g. for 3.12\% of supervision $\lambda = 200$ and for 12.5\% supervision $\lambda = 50$. This is due to the fact that in these cases only 3.12\% / 12.5\% of labeled instances contribute to $L_{att}$, while all instances contribute to $L_{rec}$.

When integrating 3.12\% of the available annotated data into the model we significantly improve the accuracy from 24.66\% to 33.02\% (VGG-CLS) and from 28.94\% to 42.32\% (VGG-DET). The accuracy further increases when providing more annotations,  reaching 42.43\% for VGG-CLS and 48.38\% for VGG-DET when using all annotations.
As ablation of our semi-supervised model we evaluated the supervised model while only using the respective $x$\% of annotated data. We observed consistent improvement of our semi-supervised model over the supervised model. %
Intrestingly, when using all available supervision, $L_{rec}$ still helps to improve performance over the  supervised model (42.43\% vs. 41.56\%, 48.38\% vs. 47.81\%). Our intuition for this is that $L_{att}$ only has a single correct bounding box (which overlaps most with the ground truth), while $L_{rec}$ can also learn from overlapping boxes with high but not best overlap.

\newcommand{\midruleValLong}{\cmidrule(rr){1-1} \cmidrule(rr){2-9} \cmidrule(rr){10-10}}
\begin{table}[t]
\scriptsize
\center
\begin{tabular}{lccccccccc}
\toprule
Phrase type & peo- & clo- & body- & ani- & vehi- & instru- & scene & other & novel \\
  & ple & thing & parts & mals & cles & ments &  &  &  \\
\midruleValLong
Number of instances & 5,656 & 2,306 & 523 & 518 & 400 & 162 & 1,619 & 3,374 & 2,214 \\
\midruleValLong
\multicolumn{10}{l}{\textbf{Unsupervised training}} \\
GroundeR (\tiny{VGG-DET}) & 44.32 & 9.02 & 0.96 & 46.91 & 46.00 & 19.14 & 28.23 & 16.98 & 25.43 \\
\midruleValLong
\multicolumn{10}{l}{\textbf{Supervised training}} \\
CCA embedding \cite{plummer15iccv} & 29.58 & 24.20 & 10.52 & 33.40 & 34.75 & 35.80 & 20.20 & 20.75 & n/a \\
GroundeR (\tiny{VGG-CLS}) & 53.80 & 34.04 & 7.27 & 49.23 & 58.75 & 22.84 & 52.07 & 24.13 & 34.28 \\
GroundeR (\tiny{VGG-DET}) & 61.00 & 38.12 & 10.33 & 62.55 & 68.75 & 36.42 & 58.18 & 29.08 & 40.83 \\
\midruleValLong
\multicolumn{2}{l}{\textbf{Semi-supervised training}} \\
GroundeR (\tiny{VGG-DET}) \scriptsize{3.12\% annot.} & 56.51 & 29.84 & 9.18 & 57.34 & 59.75 & 28.40 & 50.71 & 24.48 & 34.28 \\
GroundeR (\tiny{VGG-DET}) \scriptsize{100.0\% annot.} & 60.24 & 39.16 & 14.34 & 64.48 & 67.50 & 38.27 & 59.17 & 30.56 & 42.37 \\
\midruleValLong
Proposal upperbound & 85.93 & 66.70 & 41.30 & 84.94 & 89.00 & 70.99 & 91.17 & 69.29 & 79.90 \\
\bottomrule\\
\end{tabular}
\caption{Detailed phrase localization, Flickr 30k Entities, accuracy in \%.}
\label{tbl:testset}
\end{table}

\myparagraph{Results per phrase type.}
Flickr 30k Entities dataset provides a ``type of phrase'' annotation for each phrase, which we analyze in Table \ref{tbl:testset}. Our unsupervised approach does well on phrases like ``people'', ``animals'', ``vehicles'' and worse on ``clothing'' and ``body parts''. This could be due to confusion between people and their clothing or body parts. To address this, one could jointly model the phrases and add spatial relations between them in the model. Body parts are also the most challenging type to detect, with the proposal upper-bound of only $41.3\%$. The supervised model with VGG-CLS features outperforms \cite{plummer15iccv} in all  types except ``body parts'' and ``instruments'', while with VGG-DET it is better or similar in all types. Semi-supervised model brings further significant performance improvements, in particular for ``body parts''.
In the last column we report the accuracy for novel phrases, \ie the ones which did not appear in the training data. %
On these phrases our approach maintains high performance, although it is lower than the overall accuracy. This shows that learned language representation is effective and allows transfer to unseen phrases.

\myparagraph{Summary Flickr 30k Entities.}
Our unsupervised approach performs similar (VGG-CLS) or better (VGG-DET) than the fully supervised methods of \cite{plummer15iccv} and \cite{hu16cvpr} (\tableref{tbl:testset_flickr}). Incorporating a small amount of supervision (\eg 3.12\% of annotated data) allows us to outperform \cite{plummer15iccv} and \cite{hu16cvpr} also when VGG-CLS features are used. Our best supervised model achieves 47.81\%, surpassing all the previously reported results, including \cite{wang2016cvpr}. Our semi-supervised model efficiently exploits the reconstruction loss $L_{rec}$ which allows it to outperform  the supervised model.

\begin{table*}[t]
\scriptsize
\center
\begin{tabular}{lccc}
\toprule
Approach & \multicolumn{3}{c}{Accuracy} \\
         & Other & VGG & VGG+SPAT \\
\midruleValShort
\multicolumn{4}{l}{\textbf{Unsupervised training}} \\
LRCN \cite{donahue15cvpr} (reported in \cite{hu16cvpr}) & 8.59 & - & - \\
CAFFE-7K \cite{guadarrama2014open} (reported in \cite{hu16cvpr}) & 10.38 & - & - \\
GroundeR & - & 10.69 & 10.70 \\
\midruleValShort
\multicolumn{4}{l}{\textbf{Supervised training}} \\
SCRC \cite{hu16cvpr} & - & - & 17.93 \\
GroundeR & - & 23.44 & 26.93 \\
\midruleValShort
\multicolumn{4}{l}{\textbf{Semi-supervised training}} \\
GroundeR 3.12\% annot. & - & 13.70 & 15.03 \\
GroundeR 6.25\% annot. & - & 16.19 & 19.53 \\
GroundeR 12.5\% annot. & - & 19.02 & 21.65 \\
GroundeR 25.0\% annot. & - & 21.43 & 24.55 \\
GroundeR 50.0\% annot. & - & 22.67 & 25.51 \\
GroundeR 100.0\% annot. & - & 24.18 & 28.51 \\
\midruleValShort
Proposal upperbound & 59.38 & 59.38 & 59.38 \\
\bottomrule\\
\end{tabular}
\caption{Phrase localization performance on ReferItGame with different levels of bounding box supervision, accuracy in \%.}
\label{tbl:testset_referit}
\end{table*}

\subsection{Experiments on ReferItGame dataset}
Table \ref{tbl:testset_referit} summarizes results on the ReferItGame dataset. We compare our approach to the previously introduced fully supervised method SCRC \cite{hu16cvpr}, as well as provide reference numbers for two other baselines: LRCN \cite{donahue15cvpr} and CAFFE-7K \cite{guadarrama2014open} reported in \cite{hu16cvpr}. The LRCN baseline of \cite{hu16cvpr} is using the image captioning model LRCN \cite{donahue15cvpr} trained on MSCOCO \cite{coco2014} to score how likely the query phrase is to be generated for the proposal box. CAFFE-7K is a large scale object classifier trained on ImageNet \cite{deng09cvpr} to distinguish 7K classes. \cite{guadarrama2014open} predicts a class for each proposal box and constructs a word bag with all the synonyms of the class-name based on WordNet \cite{fellbaum:wordnet}. The obtained word bag is then compared to the query phrase after both are projected to a joint vector space. Both approaches are unsupervised w.r.t. the phrase bounding box annotations. Table \ref{tbl:testset_referit} reports the results of our approach with VGG, as well as VGG+SPAT features of \cite{hu16cvpr}.

\myparagraph{Unsupervised training.}
In the unsupervised scenario our GroundeR performs competitive with the LRCN and CAFFE-7K baselines, achieving 10.7\% accuracy. We note that in this case VGG and VGG+SPAT perform similarly.

\myparagraph{Supervised training.}
In the supervised scenario we compare to the best prior work on this dataset, SCRC \cite{hu16cvpr}, which reaches 17.93\% accuracy. Our supervised approach, which uses identical visual features, significantly improves this performance to 26.93\%. %

\myparagraph{Semi-supervised training.}
Moving to the semi-supervised scenario again demonstrates performance improvements, similar to the ones observed on Flickr 30k Entities datset. Even the small amount of supervision (3.12\%) significantly improves performance to 15.03\% (VGG+SPAT), while with 100\% of annotations we achieve 28.51\%, outperforming the supervised model.  

\myparagraph{Summary ReferItGame dataset.}
While the unsupervised model only slightly improves over prior work, the semi-supervised version can effectively learn from few labeled training instances, and with all supervision it achieves 28.51\%, improving over \cite{hu16cvpr} by a large margin of 10.6\%. Overall the performance on ReferItGame dataset is significantly lower than on Flickr 30k Entities. We attribute this to two facts. First, the training set of ReferItGame is rather small compared to Flickr 30k (9k vs. 29k images). Second, the proposal upperbound on ReferItGame is significantly lower than on Flickr 30k Entities (59.38\% vs 77.90\%) due to the complex nature of the described objects and ``stuff" image regions.

\subsection{Qualitative results}

\begin{figure*}[t]
\center
\begin{tabular}{c@{\ \ \ \ \ \ }c@{\ \ \ \ }c}
\includegraphics[clip=true,width=0.28\textwidth,height=0.16\textheight,keepaspectratio]{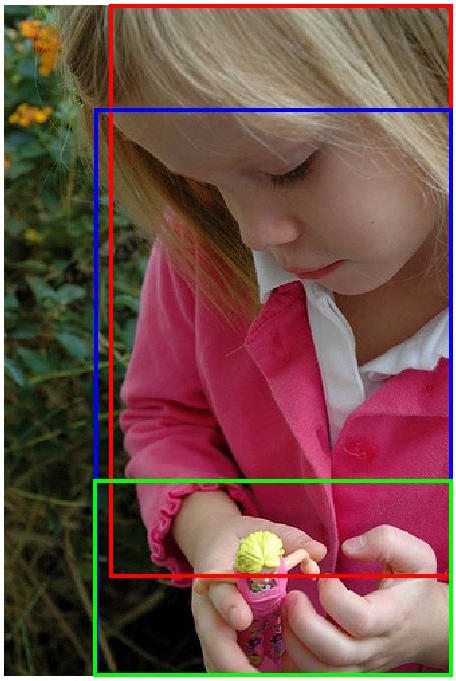} & 
\includegraphics[clip=true,width=0.28\textwidth,height=0.16\textheight,keepaspectratio]{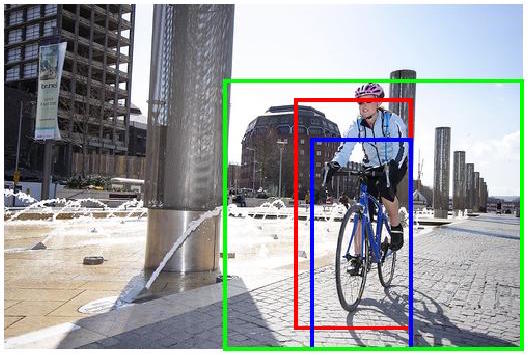} & 
\includegraphics[clip=true,width=0.28\textwidth,height=0.16\textheight,keepaspectratio]{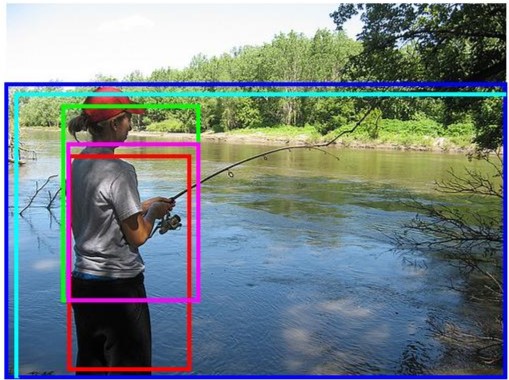} \\
\includegraphics[clip=true,width=0.28\textwidth,height=0.16\textheight,keepaspectratio]{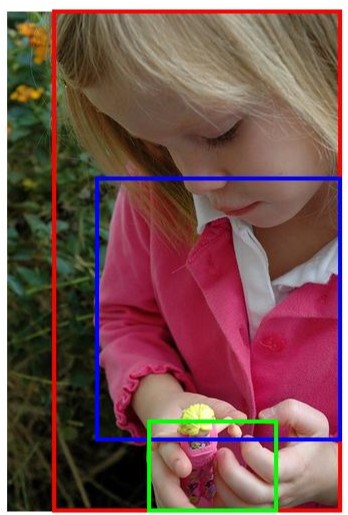} & 
\includegraphics[clip=true,width=0.28\textwidth,height=0.16\textheight,keepaspectratio]{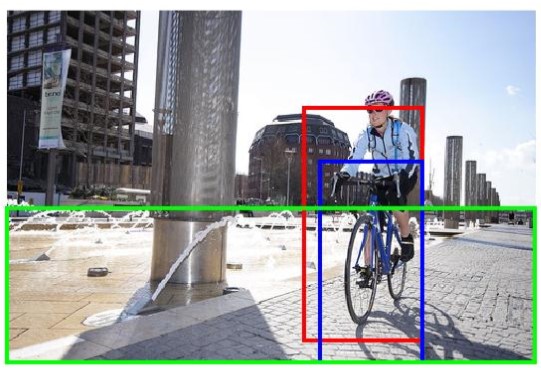} & 
\includegraphics[clip=true,width=0.28\textwidth,height=0.16\textheight,keepaspectratio]{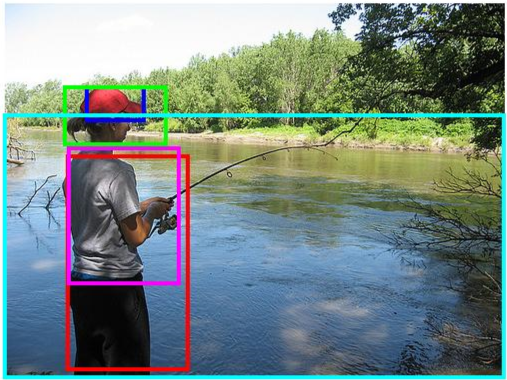} \\
\multicolumn{1}{m{2.5cm}}{\small{
\textcolor{red}{A little girl} in \textcolor{blue}{a pink shirt} is looking at \textcolor{green}{a toy doll}.\newline ~
}} &
\multicolumn{1}{m{3.5cm}}{\small{
\textcolor{red}{A woman} is riding \textcolor{blue}{a bicycle} on \textcolor{green}{the pavement}.\newline
}} & 
\multicolumn{1}{m{3.7cm}}{\small{
\textcolor{red}{A girl} with \textcolor{blue}{a red cap}, \textcolor{green}{hair} tied up and \textcolor{magenta}{a gray shirt} is fishing in \textcolor{cyan}{a calm lake}.
}} 
\end{tabular}
\caption{Qualitative results on the test set of Flickr 30k Entities. Top : GroundeR (VGG-DET) unsupervised, bottom: GroundeR (VGG-DET) supervised.}
\label{fig:qualitative_flickr}
\end{figure*}

\begin{figure*}[t]
\center
\begin{tabular}{c@{\ \ }c@{\ \ }c}
\includegraphics[clip=true,width=0.28\textwidth,height=0.16\textheight,keepaspectratio]{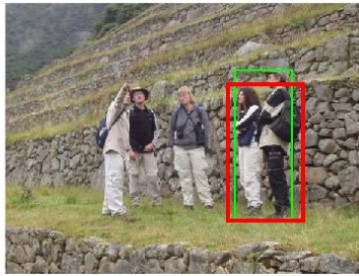} & 
\includegraphics[clip=true,width=0.28\textwidth,height=0.16\textheight,keepaspectratio]{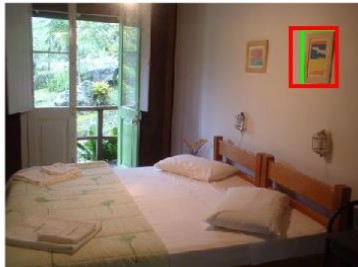} &
\includegraphics[clip=true,width=0.28\textwidth,height=0.16\textheight,keepaspectratio]{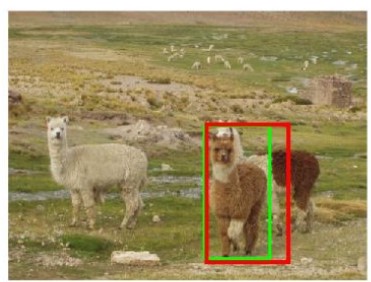} \\
\multicolumn{1}{m{3.5cm}}{\small{
two people on right}} &
\multicolumn{1}{m{3.5cm}}{\small{
picture of a bird flying above sand}} &
\multicolumn{1}{m{3.5cm}}{\small{
dat alpaca up in front, total coffeelate swag}} \\
\includegraphics[clip=true,width=0.28\textwidth,height=0.16\textheight,keepaspectratio]{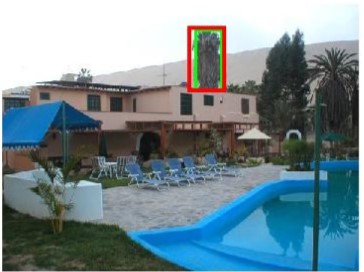} &
\includegraphics[clip=true,width=0.28\textwidth,height=0.16\textheight,keepaspectratio]{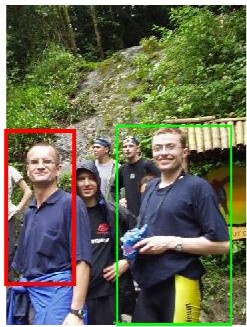} & 
\includegraphics[clip=true,width=0.28\textwidth,height=0.16\textheight,keepaspectratio]{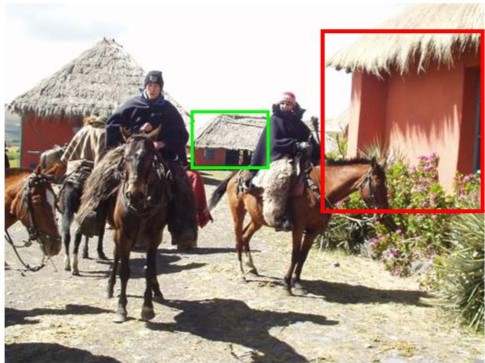} \\
\multicolumn{1}{m{3.5cm}}{\small{
palm tree coming out of the top of the building}} &
\multicolumn{1}{m{3.5cm}}{\small{
guy with blue shirt and yellow shorts}} &
\multicolumn{1}{m{3.5cm}}{\small{
hut to the nearest left of the person on the right}} \\
\end{tabular}
\caption{Qualitative results on the test set of ReferItGame: GroundeR (VGG+SPAT) supervised. Green: ground-truth box, red: predicted box.}
\label{fig:qualitative_referit}
\end{figure*}

We provide qualitative results on Flickr 30K Entities dataset in Figure \ref{fig:qualitative_flickr}. We compare our unsupervised and supervised approaches, both with VGG-DET features. The supervised approach visibly improves the localization quality over the unsupervised approach, which nevertheless is able to localize many phrases correctly.
Figure \ref{fig:qualitative_referit} presents qualitative results on ReferItGame dataset. We show the predictions of our supervised approach, as well as the ground-truth boxes. One can see the difficulty of the task from the presented examples, including two failures in the bottom row. One requires good language understanding in order to correctly ground such complex phrases. In order to ground expressions like ``hut to the nearest left of the person on the right'' we would need to additionally model relations between objects, an interesting direction for future work.

\section{Conclusion}

In this work we address the challenging task of grounding unconstrained natural phrases in images. We consider different scenarios of available bounding box supervision at training time, namely none, little, and full supervision. We propose a novel approach, \approach, which learns to localize phrases in images by attending to the correct box proposal and reconstructing the phrase and is able to operate in all of these supervision scenarios. 
In the unsupervised scenario we are competitive or better than related work. %
Our semi-supervised approach works well with  a small portion of available annotated data and takes advantage of the unsupervised data to outperform purely supervised training using the same amount of labeled data. %
It outperforms  state-of-the-art, both on Flickr 30k Entities and ReferItGame dataset, by 4.5\% and 10.6\%, respectively.

Our approach is rather general and it could be applied to other regions such as segmentation proposals instead of bounding box proposals. Possible extensions are to include constraints within sentences at training time, jointly reason about multiple phrases, and to take into account spatial relations between them.

\paragraph{Acknowledgements.}
Marcus Rohrbach was supported by a fellowship within the FITweltweit-Program of the German Academic Exchange Service (DAAD). This work was supported by DARPA, AFRL, DoD MURI award N000141110688, NSF awards IIS-1427425 and IIS-1212798, and the Berkeley Artificial Intelligence Research (BAIR) Lab.

\bibliographystyle{splncs03}
\bibliography{biblioLong,rohrbach16eccv}

\begin{thebibliography}{10}
\providecommand{\url}[1]{\texttt{#1}}
\providecommand{\urlprefix}{URL }

\bibitem{ammar2014conditional}
Ammar, W., Dyer, C., Smith, N.A.: Conditional random field autoencoders for
  unsupervised structured prediction. In: Advances in Neural Information
  Processing Systems (NIPS) (2014)

\bibitem{bahdanau2014neural}
Bahdanau, D., Cho, K., Bengio, Y.: Neural machine translation by jointly
  learning to align and translate. In: International Conference on Learning
  Representations (ICLR) (2015)

\bibitem{blaschko2010simultaneous}
Blaschko, M., Vedaldi, A., Zisserman, A.: Simultaneous object detection and
  ranking with weak supervision. In: Advances in Neural Information Processing
  Systems (NIPS). pp. 235--243 (2010)

\bibitem{chen2015webly}
Chen, X., Gupta, A.: Webly supervised learning of convolutional networks. In:
  Proceedings of the IEEE International Conference on Computer Vision (ICCV)
  (2015)

\bibitem{chen15cvpr}
Chen, X., Zitnick, C.L.: Mind's eye: A recurrent visual representation for
  image caption generation. Proceedings of the IEEE Conference on Computer
  Vision and Pattern Recognition (CVPR)  (2015)

\bibitem{cinbis2014multi}
Cinbis, R.G., Verbeek, J., Schmid, C.: Multi-fold mil training for weakly
  supervised object localization. In: Proceedings of the IEEE Conference on
  Computer Vision and Pattern Recognition (CVPR) (2014)

\bibitem{deng09cvpr}
Deng, J., Dong, W., Socher, R., Li, L.J., Li, K., Fei-Fei, L.: Imagenet: A
  large-scale hierarchical image database. In: Proceedings of the IEEE
  Conference on Computer Vision and Pattern Recognition (CVPR) (2009)

\bibitem{divvala2014learning}
Divvala, S., Farhadi, A., Guestrin, C.: Learning everything about anything:
  Webly-supervised visual concept learning. In: Proceedings of the IEEE
  Conference on Computer Vision and Pattern Recognition (CVPR) (2014)

\bibitem{donahue15cvpr}
Donahue, J., Hendricks, L.A., Guadarrama, S., Rohrbach, M., Venugopalan, S.,
  Saenko, K., Darrell, T.: Long-term recurrent convolutional networks for
  visual recognition and description. In: Proceedings of the IEEE Conference on
  Computer Vision and Pattern Recognition (CVPR) (2015)

\bibitem{everingham2010pascal}
Everingham, M., Van~Gool, L., Williams, C.K., Winn, J., Zisserman, A.: The
  pascal visual object classes (voc) challenge. International Journal of
  Computer Vision (IJCV)  88(2),  303--338 (2010)

\bibitem{fellbaum:wordnet}
Fellbaum, C.: WordNet: An Electronical Lexical Database. The MIT Press (1998)

\bibitem{girshick2015fast}
Girshick, R.: Fast {R-CNN}. In: Proceedings of the IEEE International
  Conference on Computer Vision (ICCV) (2015)

\bibitem{glorot2010understanding}
Glorot, X., Bengio, Y.: Understanding the difficulty of training deep
  feedforward neural networks. In: International conference on artificial
  intelligence and statistics. pp. 249--256 (2010)

\bibitem{gong2014eccv}
Gong, Y., Wang, L., Hodosh, M., Hockenmaier, J., Lazebnik, S.: Improving
  image-sentence embeddings using large weakly annotated photo collections. In:
  Proceedings of the European Conference on Computer Vision (ECCV), pp.
  529--545. Springer (2014)

\bibitem{guadarrama2014open}
Guadarrama, S., Rodner, E., Saenko, K., Zhang, N., Farrell, R., Donahue, J.,
  Darrell, T.: Open-vocabulary object retrieval. In: Robotics: science and
  systems (2014)

\bibitem{he2015delving}
He, K., Zhang, X., Ren, S., Sun, J.: Delving deep into rectifiers: Surpassing
  human-level performance on imagenet classification. In: Proceedings of the
  IEEE Conference on Computer Vision and Pattern Recognition (CVPR) (2015)

\bibitem{hochreiter1997long}
Hochreiter, S., Schmidhuber, J.: Long short-term memory. Neural computation
  9(8),  1735--1780 (1997)

\bibitem{hu16cvpr}
Hu, R., Xu, H., Rohrbach, M., Feng, J., Saenko, K., Darrell, T.: Natural
  language object retrieval. In: Proceedings of the IEEE Conference on Computer
  Vision and Pattern Recognition (CVPR) (2016)

\bibitem{ioffe2015batch}
Ioffe, S., Szegedy, C.: Batch normalization: Accelerating deep network training
  by reducing internal covariate shift. arXiv:1502.03167  (2015)

\bibitem{jia2014caffe}
Jia, Y., Shelhamer, E., Donahue, J., Karayev, S., Long, J., Girshick, R.,
  Guadarrama, S., Darrell, T.: Caffe: Convolutional architecture for fast
  feature embedding. In: Proceedings of the ACM International Conference on
  Multimedia. pp. 675--678. ACM (2014)

\bibitem{jin2015aligning}
Jin, J., Fu, K., Cui, R., Sha, F., Zhang, C.: Aligning where to see and what to
  tell: image caption with region-based attention and scene factorization.
  arXiv:1506.06272  (2015)

\bibitem{johnson2015cvpr}
Johnson, J., Krishna, R., Stark, M., Li, L.J., Shamma, D., Bernstein, M.,
  Fei-Fei, L.: Image retrieval using scene graphs. In: Proceedings of the IEEE
  Conference on Computer Vision and Pattern Recognition (CVPR). pp. 3668--3678
  (2015)

\bibitem{joulin2014eccv}
Joulin, A., Tang, K., Fei-Fei, L.: Efficient image and video co-localization
  with frank-wolfe algorithm. In: Proceedings of the European Conference on
  Computer Vision (ECCV) (2014)

\bibitem{karpathy15cvpr}
Karpathy, A., Fei-Fei, L.: Deep visual-semantic alignments for generating image
  descriptions. In: Proceedings of the IEEE Conference on Computer Vision and
  Pattern Recognition (CVPR) (2015)

\bibitem{karpathy14nips}
Karpathy, A., Joulin, A., Fei-Fei, L.: Deep fragment embeddings for
  bidirectional image sentence mapping. In: Advances in Neural Information
  Processing Systems (NIPS) (2014)

\bibitem{kazemzadeh14emnlp}
Kazemzadeh, S., Ordonez, V., Matten, M., Berg, T.L.: Referit game: Referring to
  objects in photographs of natural scenes. In: Proceedings of the Conference
  on Empirical Methods in Natural Language Processing (EMNLP) (2014)

\bibitem{kingma2014adam}
Kingma, D., Ba, J.: Adam: A method for stochastic optimization. arXiv:1412.6980
   (2014)

\bibitem{kong14cvpr}
Kong, C., Lin, D., Bansal, M., Urtasun, R., Fidler, S.: What are you talking
  about? text-to-image coreference. In: Proceedings of the IEEE Conference on
  Computer Vision and Pattern Recognition (CVPR). pp. 3558--3565. IEEE (2014)

\bibitem{krishnamurthy13tacl}
Krishnamurthy, J., Kollar, T.: Jointly learning to parse and perceive:
  connecting natural language to the physical world. Transactions of the
  Association for Computational Linguistics (TACL)  (2013)

\bibitem{kwak15arxiv}
Kwak, S., Cho, M., Laptev, I., Ponce, J., Schmid, C.: Unsupervised object
  discovery and tracking in video collections. In: Proceedings of the IEEE
  International Conference on Computer Vision (ICCV) (2015)

\bibitem{lin14cvpr}
Lin, D., Fidler, S., Kong, C., Urtasun, R.: Visual semantic search: Retrieving
  videos via complex textual queries. In: Proceedings of the IEEE Conference on
  Computer Vision and Pattern Recognition (CVPR). pp. 2657--2664. IEEE (2014)

\bibitem{coco2014}
Lin, T.Y., Maire, M., Belongie, S., Hays, J., Perona, P., Ramanan, D.,
  Doll{\'a}r, P., Zitnick, C.L.: Microsoft coco: Common objects in context. In:
  Proceedings of the European Conference on Computer Vision (ECCV) (2014)

\bibitem{mao16cvpr}
Mao, J., Huang, J., Toshev, A., Camburu, O., Yuille, A., Murphy, K.: Generation
  and comprehension of unambiguous object descriptions. In: Proceedings of the
  IEEE Conference on Computer Vision and Pattern Recognition (CVPR) (2016)

\bibitem{matuszek12icml}
Matuszek, C., Fitzgerald, N., Zettlemoyer, L., Bo, L., Fox, D.: A joint model
  of language and perception for grounded attribute learning. In: Proceedings
  of the International Conference on Machine Learning (ICML) (2012)

\bibitem{plummer15iccv}
Plummer, B., Wang, L., Cervantes, C., Caicedo, J., Hockenmaier, J., Lazebnik,
  S.: Flickr30k entities: Collecting region-to-phrase correspondences for
  richer image-to-sentence models. In: Proceedings of the IEEE International
  Conference on Computer Vision (ICCV) (2015)

\bibitem{sadeghi2015viske}
Sadeghi, F., Divvala, S.K., Farhadi, A.: Viske: Visual knowledge extraction and
  question answering by visual verification of relation phrases. In:
  Proceedings of the IEEE Conference on Computer Vision and Pattern Recognition
  (CVPR) (2015)

\bibitem{simonyan2014very}
Simonyan, K., Zisserman, A.: Very deep convolutional networks for large-scale
  image recognition. In: International Conference on Learning Representations
  (ICLR) (2015)

\bibitem{song2014learning}
Song, H.O., Girshick, R., Jegelka, S., Mairal, J., Harchaoui, Z., Darrell, T.:
  On learning to localize objects with minimal supervision. arXiv:1403.1024
  (2014)

\bibitem{sutskever14nips}
Sutskever, I., Vinyals, O., Le, Q.V.: Sequence to sequence learning with neural
  networks. In: Advances in Neural Information Processing Systems (NIPS). pp.
  3104--3112 (2014)

\bibitem{tang2014cvpr}
Tang, K., Joulin, A., Li, L.J., Fei-Fei, L.: Co-localization in real-world
  images. In: Proceedings of the IEEE Conference on Computer Vision and Pattern
  Recognition (CVPR). IEEE (2014)

\bibitem{tapaswi2015cvpr}
Tapaswi, M., B{\"a}uml, M., Stiefelhagen, R.: Book2movie: Aligning video scenes
  with book chapters. In: Proceedings of the IEEE Conference on Computer Vision
  and Pattern Recognition (CVPR). pp. 1827--1835 (2015)

\bibitem{uijlings2013selective}
Uijlings, J.R., van~de Sande, K.E., Gevers, T., Smeulders, A.W.: Selective
  search for object recognition. International Journal of Computer Vision
  (IJCV)  104(2) (2013)

\bibitem{vincent2008extracting}
Vincent, P., Larochelle, H., Bengio, Y., Manzagol, P.A.: Extracting and
  composing robust features with denoising autoencoders. In: Proceedings of the
  International Conference on Machine Learning (ICML) (2008)

\bibitem{vinyals14arxiv}
Vinyals, O., Toshev, A., Bengio, S., Erhan, D.: Show and tell: {A} neural image
  caption generator. Proceedings of the IEEE Conference on Computer Vision and
  Pattern Recognition (CVPR)  (2015)

\bibitem{wang2016cvpr}
Wang, L., Li, Y., Lazebnik, S.: Learning deep structure-preserving image-text
  embeddings. In: Proceedings of the IEEE Conference on Computer Vision and
  Pattern Recognition (CVPR) (2016)

\bibitem{xu2015arxiv}
Xu, K., Ba, J., Kiros, R., Courville, A., Salakhutdinov, R., Zemel, R., Bengio,
  Y.: Show, attend and tell: Neural image caption generation with visual
  attention. In: Proceedings of the International Conference on Machine
  Learning (ICML) (2015)

\bibitem{yao2015iccv}
Yao, L., Torabi, A., Cho, K., Ballas, N., Pal, C., Larochelle, H., Courville,
  A.: Describing videos by exploiting temporal structure. In: Proceedings of
  the IEEE International Conference on Computer Vision (ICCV) (2015)

\bibitem{yeung2015every}
Yeung, S., Russakovsky, O., Jin, N., Andriluka, M., Mori, G., Fei-Fei, L.:
  Every moment counts: Dense detailed labeling of actions in complex videos.
  arXiv:1507.05738  (2015)

\bibitem{young2014image}
Young, P., Lai, A., Hodosh, M., Hockenmaier, J.: From image descriptions to
  visual denotations: New similarity metrics for semantic inference over event
  descriptions. Transactions of the Association for Computational Linguistics
  2,  67--78 (2014)

\bibitem{yu2013acl}
Yu, H., Siskind, J.M.: Grounded language learning from video described with
  sentences. In: Proceedings of the Annual Meeting of the Association for
  Computational Linguistics (ACL). pp. 53--63 (2013)

\bibitem{yu15arxiv}
Yu, H., Siskind, J.M.: Sentence directed video object codetection.
  arXiv:1506.02059  (2015)

\bibitem{zhu2015aligning}
Zhu, Y., Kiros, R., Zemel, R., Salakhutdinov, R., Urtasun, R., Torralba, A.,
  Fidler, S.: Aligning books and movies: Towards story-like visual explanations
  by watching movies and reading books. In: Proceedings of the IEEE
  International Conference on Computer Vision (ICCV) (2015)

\bibitem{zitnick2014eccv}
Zitnick, C.L., Doll{\'a}r, P.: Edge boxes: Locating object proposals from
  edges. In: Computer Vision--ECCV 2014. pp. 391--405. Springer (2014)

\end{thebibliography}

\end{document}